\documentclass[letterpaper]{article}
\usepackage[latin9]{inputenc}
\usepackage{float}
\usepackage{amsmath}
\usepackage{graphicx}

\makeatletter

\providecommand{\tabularnewline}{\\}
\floatstyle{ruled}
\newfloat{algorithm}{tbp}{loa}
\providecommand{\algorithmname}{Algorithm}
\floatname{algorithm}{\protect\algorithmname}

\usepackage{ijcai13}

\usepackage{times}
\title{Learning Visual Symbols for Parsing Human Poses in Images}

\author{Fang Wang$^{1,2}$ $~$ $\qquad$ $~$ Yi Li$^{2}$\\
$^{1}$ Nanjing University of Science and Technology, China\\
$^{2}$ NICTA, Canberra, Australia \\
$\{$fang.wang, yi.li$\}$@nicta.com.au}

\makeatother

\begin{document}
\maketitle
\begin{abstract}
Parsing human poses in images is fundamental in extracting critical
visual information for artificial intelligent agents. Our goal is
to learn self-contained body part representations from images, which
we call \textit{visual symbols}, and their \textit{symbol-wise} geometric
contexts in this parsing process. Each symbol is individually learned
by categorizing visual features leveraged by geometric information.
In the categorization, we use Latent Support Vector Machine followed
by an efficient cross validation procedure to learn visual symbols. Then, these symbols naturally
define geometric contexts of body parts in a fine granularity. When the structure of the compositional parts
is a tree, we derive an efficient approach to estimating human poses in images.
Experiments on two large datasets suggest our approach outperforms
state of the art methods.
\end{abstract}

\section{Introduction}

Parsing human poses in images has been a classical topic in artificial
intelligence for decades \cite{Marr:1982:VCI:1095712}. This research
facilitates a number of fundamental studies ranging from visual perception
\cite{10.1371/journal.pone.0035757} to computer vision \cite{Felzenszwalb:2005:PSO:1024426.1024429},
to particularly cognitive robotics in recent years \cite{DBLP:journals/ijhr/JenkinsSL07}.

We focus on learning visual representations of body parts in this
parsing process, which we call \textit{visual symbols}. \cite{Marr:1982:VCI:1095712}
has already argued that any meaningful representation of the human
body should be \textit{self-contained} in a semantic hierarchy. 
In his work, the main ingredients of ``self-containedness" are (i) self-contained unit must have a limited complexity, such that (ii) information appears in geometric contexts appropriate for recognition, and (iii) the representation can be processed flexibly.

However, recent research is contrary to this intuitive philosophy.
The body parts are frequently represented by plain cardboard models
(\textit{e.g., }joints or limbs only). Since each part is not distinctive,
the visual units are considered as an approach for computing probability
of the body part locations, and the geometric contexts are coarsely
defined as simple distributions between parts. As such, the inference
models may have to go beyond tree structures to model long range interactions
in this coarse structure (\textit{e.g.}, \cite{sun2012efficient}).
This essentially makes the problem less tractable, and approximate
inference has to be adopted. 

\begin{figure*}[!t]
\begin{centering}
\includegraphics[width=0.6\textheight]{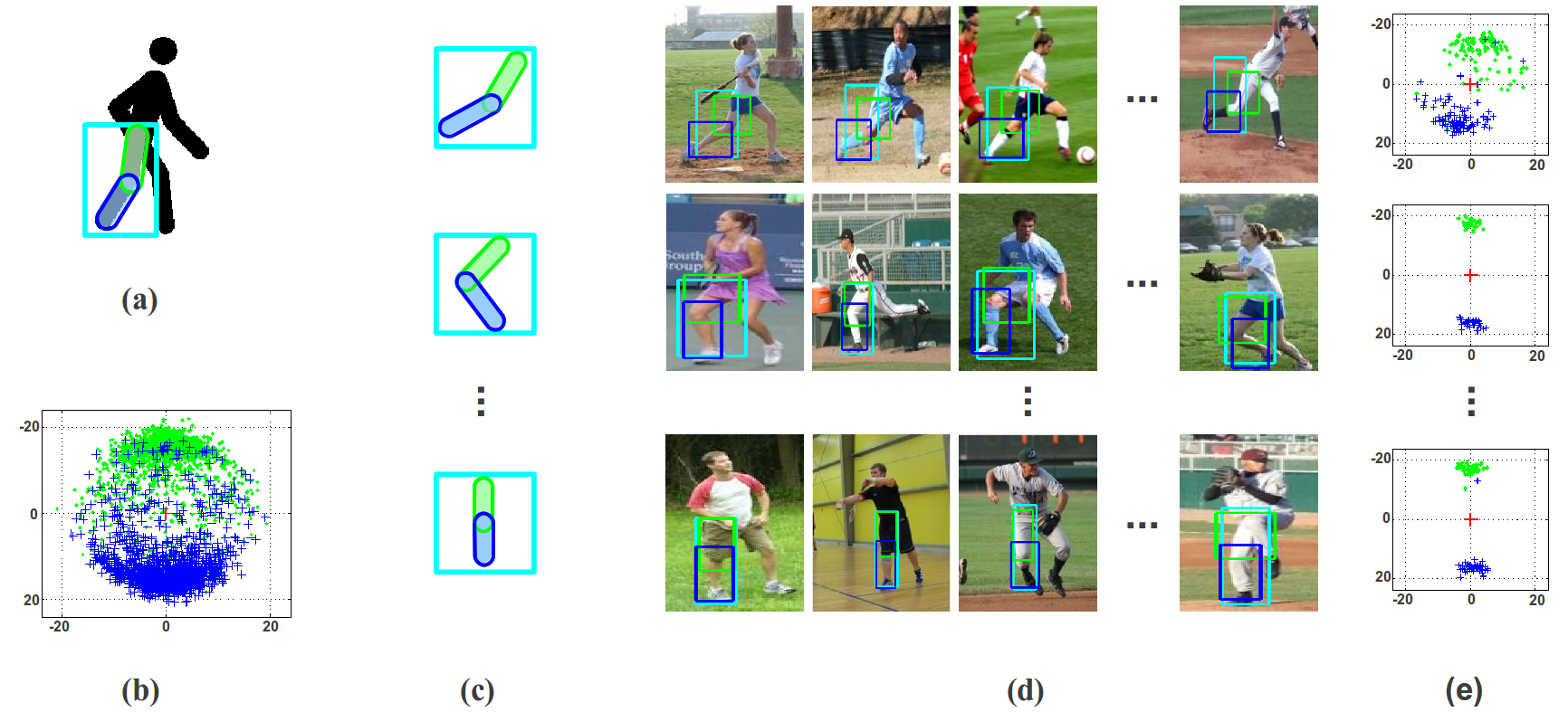}
\par\end{centering}

\caption{\label{fig:illustration}Our motivation. (a) human body. Green denotes
upper leg, Blue denotes lower leg. (b) relative distances between
upper/lower leg and leg in a large dataset, respectively. If we can
group the instances to symbols (c)(d), we can easily see that relative
distances can be modeled in a fine scale (e). The coordinate is defined
in the image space (pixels).}
\end{figure*}

Can we still use exact and fast inference to remedy the problems caused
by the deformable nature of human beings? We propose to use compositional
parts and exploit \textit{symbol-wise geometric contexts} map for
effective pose parsing in still images. Our part representation is
compositional, \textit{i.e.}, each part may contain one or more physical
joints of the human body, and the relationship between two parts can
be either hierarchical (parent-child, \textit{e.g.}, leg and upper leg, Fig.
\ref{fig:illustration}) or flat. This allows us to categorize distinctive
visual features for body parts, which are the descriptors that characterize the properties of image patches, and eventually model pairwise interactions in fine granularity.

A popular view of the geometric context is that pairwise relationship between two parts
can be characterized by a distribution. Fig. \ref{fig:illustration}b
shows a distribution of relative locations between upper/lower leg
(denoted by green/blue colors) and ``leg'' (denoted by cyan), respectively.

It is seemingly legitimate to assume both point sets satisfy normal
distributions, but let us take a further look at the data. Assuming
we have learned the symbols of the ``leg'' as depicted in Fig. \ref{fig:illustration}c.
We can group all the instances in each point set to a few categories
(Fig. \ref{fig:illustration}c and \ref{fig:illustration}d). Further,
we redraw the relative locations of upper/lower leg that are only
associated to the corresponding symbols of the leg (Fig. \ref{fig:illustration}e).
Clearly, we have two observations: 1) relative locations may have
different distributions and exhibit different characteristics, and
2) some categories may have similar distributions, but they are compactly
distributed and have much smaller variances compared to Fig. \ref{fig:illustration}b.
Therefore, it is much easier to model these subsets separately. 

The concept of symbols, although it is not new in vision and artificial intelligence, has not been used explicitly in many state of the art pose estimation methods.
Such a set of learned symbols enables encoding the symbol-wise geometric
information in a finer scale, and provides more information in inference.
Therefore, it is critical to learn symbols for compositional
parts. 

With the help of geometric information, we categorize visual
features from part instances, and use cross validation to select the
best categories. 
We used Histogram of Oriented Gradient (HOG) \cite{dalal2005histograms} in this paper.
HOG is frequently used as the feature for the appearance model of body parts.
In this descriptor, an image is divided into smaller regions called cells, and the histogram of gradient directions of the pixels within each cell is calculated as the descriptor.

Our approach has the following contributions:
\begin{itemize}
\item We explore an effective procedure for learning self-contained symbols
of body parts in parsing poses.
\item Our symbol-wise context map naturally encodes the legitimate combinations
of human poses from images.
\item Our representation is very flexible, therefore, it is compatible with
the majority of the popular inference algorithm in human pose estimation.
\end{itemize}
Following human kinematic structure, we derive an approach to effectively
estimate human poses. We demonstrate the performance of our method
in two large datasets, and our method outperforms the state of art
methods.
\begin{figure*}[!]
\begin{centering}
\includegraphics[width=0.95\textwidth]{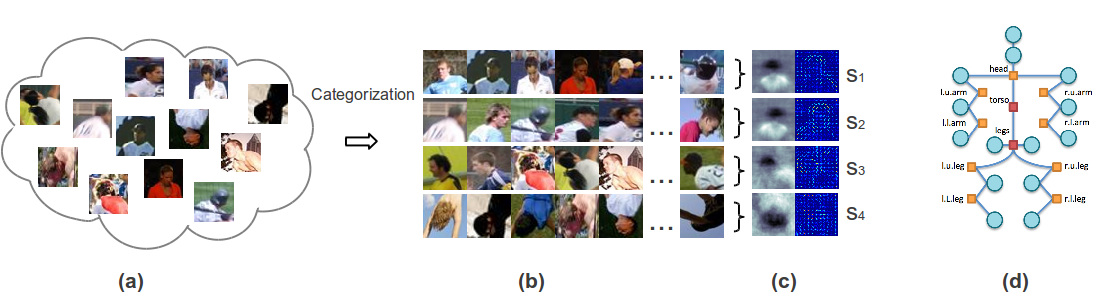}
\par\end{centering}

\caption{\label{fig:vc}Training visual symbols. Given a set of instances of
a compositional part (a), our approach categorizes these instances
(b) and summarize it to symbols by a set of linear filters (c). (d) Our tree structure model for parsing human poses. Semantically, there are two ``high level'' parts (red), nine ``mid level'' parts (orange), and 14 joints (blue) in total.}
\end{figure*}

\section{Related work}

Marr is among the first to propose a hierarchical organization of body for parsing human
poses \cite{Marr:1982:VCI:1095712}. Each model in this hierarchy
is a self-contained unit, and the geometric contexts among these units are designed for recognition. 
His structure motivated a number of approaches in computer vision and machine learning.
This idea evolves to \textit{deformable models} in recent years, where pose estimation has been formulated as a part based inference problem.

Pictorial Structure Model (PSM) \cite{Felzenszwalb:2005:PSO:1024426.1024429}
is one of the most successful deformable models. 
A tree structure graphical model is used, and pairwise terms are based on the
relative distances between corresponding body parts. 
\cite{yang2011articulated} proposed a mixtures-of-parts model for articulated pose estimation.
Instead of modeling both location and orientation of body limbs as
rigid parts (\textit{e.g.}, \cite{andriluka2009pictorial}), they used non-oriented pictorial structures with co-occurrence constraints.
Their work relies on the geometry to define clusters (called ``types" in their paper). 
Therefore, the representation is less \textit{self-contained} from Marr's point of view.

Following this research direction, \cite{sun2011articulated} used predefined symbols for a simultaneous detection of body parts and estimation of human poses. 
\cite{DBLP:conf/eccv/TianZN12} used compatibility maps in a tree structure. Latent nodes encode compatibility
between parts, and accuracy is improved because incompatible poses are pruned.
However, their ``types" variables are solely based on the geometry, and do not encode visual information.

Beyond tree structures, graphical models were proposed in pose estimation. 
\cite{tran2010improved} evaluated the performance of human parsing in full relational model.
A recent work by \cite{sun2012efficient} showed the solution of a loopy star model using Branch-and-Bound strategy.
These structures usually lead to better performance, but all require efficient approximate inference methods. 

All above methods used various definitions of card-board style parts, such as limbs \cite{andriluka2009pictorial} and joints \cite{yang2011articulated}.
This may appear easier to model, but essentially the features are less distinctive and the performance is limited by coarse geometric context.
Beyond these plain structures, researchers seek to compositional parts in recent years to characterize higher level visual representations. 

\cite{wang2011learning} proposed to use compositional parts to provide more precise
results, but his method has a higher computation cost due to the loopy graph structure.
\cite{rothrock2011human} cast human pose into AND/OR graphs, and performed human parsing using top-down scheme. 
Rich appearance models were adopted to estimate human parts. 
\cite{bourdev2010detecting} proposed to use poselets for human recognition.
Each poselet refers to a combined part that is distinctive in training images.
Please note that these poselets do not characterise geometric contexts in modeling pairwise distributions, which makes it less effective to fully capture the body dynamical structures.

\section{Approach}

Define the set of $M$ parts as $P=\{P_{i}\}$, $i\in[1,...,M]$.
One of our goals is to learn a symbol set $S$ such that an instance
of a body part can be labelled by an entry in $S$. 

In many state of the art datasets, the locations of primitive parts
(\textit{e.g.}, joints) are manually annotated. Therefore, the goal
of our approach is to learn and assign symbols for compositional
parts in the training set, and to detect compositional parts in test
images by inferring their symbols. 

We first present our approach to learning visual symbols for compositional
parts, then we derive the compatibility map for fine scale modeling
of geometric context. Finally, we adopt an efficient learning and
inference method when the structure of graphical models is a tree.

\subsection{Learning symbols for compositional parts}

Let $G=(P,E)$ denote the relationship graph, where $P$ denotes the body parts, and $E$ is the set of edges that denote the pairwise relationship between parts. An instance of a compositional part $p_{i}=(loc_{i},s_{i})$, where
$loc_{i}$ can be used for computing local geometric context (relative distance) in $G$, and $s_i$ denotes visual appearance. 

We exploit the advantages of both geometric and visual information. We first
use geometric information to coarsely group image patches for $p_{i}$
to different clusters, then we categorize the visual features in each
cluster to visual symbols. This is more computational efficient than
first categorizing visual features then grouping locations to symbols.

\subsubsection{Geometric grouping}

Given a pair of connected parts $(p_{i},p_{j})$ in $G$, we use the first part
$p_{i}$ as reference, and calculate the relative locations of $p_{j}$
with respect to $p_i$. As a result, all samples of $p_{j}$
are projected into 2D geometry space.

For efficiency consideration, we run \textit{k}-mean to group samples
in $k_{j}$ geometry clusters, each of which denotes a geometry \textit{type}
of $p_{j}$.

\subsubsection{Discovering visual categories}

Instances for compositional parts still have large variations in appearance
within a geometric type. Therefore, the geometry alone is not powerful
to characterize symbols, and we need to further learn visual categories
in each geometric group for generating more discriminative visual
symbols. 

Fig. \ref{fig:vc} illustrates the learning process. Given a set of
instances of a part (Fig. \ref{fig:vc}a), our approach categorizes
these instances to a number of subsets that are meaningful both in
geometric context and visual appearance (Fig. \ref{fig:vc}b), and
summarize it to symbols by a set of linear weights (Fig. \ref{fig:vc}c)

Let $\phi(I,p_{i})$ denotes the visual feature of $p_{i}$ in the
image $I$. For instances of $p_{i}$ within the same geometry context,
we aim at learning linear classifiers to categorize visual features. 

We followed \cite{DBLP:journals/corr/abs-1206-3714} and built a Latent
Support Vector Machine model for learning visual subcategories. Given
$N_{1}$ positive instances of a compositional part, and $N_{2}$
negative instances, we learn $K$ subcategories of this positive set.
This allows us to generate the labels $L={l_{1},l_{2},\cdots,l_{N}}$,
$l_{i}\in[1,K]$ for each instance. Our objective function is as follows
\begin{equation}
\begin{split}\arg\min_{w}\frac{1}{2}\sum_{k=1}^{K}\parallel w_{k}\parallel^{2}+C\sum_{i=1}^{N_{1}+N_{2}}\epsilon_{i},\\
y_{i}w_{l_{i}}\phi(p_{i})\geq1-\epsilon_{i},\epsilon_{i}\geq0,\\
l_{i}=\arg\max_{k}w_{k}\phi(p_{i})
\end{split}
\label{eq:1}
\end{equation}
where $y_{i}=\{1,-1\}$ denotes whether $y_{i}$ is from positive
or negative sample sets, and $w_k$ are the weights of the feature map
for each part. Similar as other clustering methods, we use \textit{k}-mean
for initializing categorization.

\subsubsection{Cross validation}

To achieve effective training when number of visual samples is small,
we turn to a cross-validation learning paradigm to discover the best
classifiers and fine tune the performance. The main idea of cross
validation is to perform a training step followed by a validation
step \cite{Singh:2012:UDM:2403006.2403013}. During the validation
stage, each classifier is evaluated on the validation set, and weak
classifiers with few detected samples will be removed from the classifier
set. 

The whole training process is conducted iteratively. Algorithm \ref{alg:Cross-validation.}
illustrates the whole training process. After cross training the ``survived''
classifiers canbe regarded as a mixture of visual symbols. 

\begin{algorithm}
\textbf{Input}: 

$\begin{array}{c}
{}\end{array}$ $H$ : training set for a compositional part $P_{i}$;

$\begin{array}{c}
{}\end{array}$ $K_{in}$ : the number of classifiers.

\textbf{Output}: 

$\begin{array}{c}
{}\end{array}$ $K_{out}$ : the number of visual categories;

$\begin{array}{c}
{}\end{array}$ $w_{k}$: linear classifiers $k=[1,...,K_{out}]$.

\textbf{Procedure}:

$\begin{array}{c}
{}\end{array}$ 1. Divide the training set equally to $H_{1}$ and $H_{2};$

$\begin{array}{c}
{}\end{array}$ 2. Train the classifiers $w_{[1,...,K_{in}]}$ on $H_{1}$ using
Eq. \ref{eq:1};

$\begin{array}{c}
{}\end{array}$ 3. Evaluate the classification result on $H_{2}$;

$\begin{array}{c}
{}\end{array}$ 4. If detected samples for $w_{i}$ is small:

$\begin{array}{c}
{}\end{array}\begin{array}{c}
{}\end{array}$$\begin{array}{c}
{}\end{array}$Remove $w_{i}$ from classifiers and $K_{in}=K_{in}-1$ ;

$\begin{array}{c}
{}\end{array}$ 5. Swap $H_{1}$ and $H_{2}$;

$\begin{array}{c}
{}\end{array}$ 6. Repeat step 1-5 for $t$ times ($t=10$ in our experiments);

$\begin{array}{c}
{}\end{array}$ 7. $K_{out}=K_{in}$ and output $w_{[1,...,K_{out}]}$ ;

\caption{Cross validation.\label{alg:Cross-validation.}}
\end{algorithm}

\subsubsection{Discussion}

The visual categorization process of a compositional part characterizes
the appearance models in a way that they can be regarded as ``templates''.
When HOG feature is used, the set of learned weights is also considered as ``HOG filters''.

In this way, our symbols encode both geometric and visual appearance information.
This makes our descriptors different from other work, because they are more discriminative and representative.

\subsection{Defining symbol-wise geometric context}

Assigning each part a symbol allows us to build a compatibility map
for any pair of symbols. Assume we have two compositional parts $p_{i}$
and $p_{j}$, each of which has symbols $s_{i}$ and $s_{j}$, respectively,
we create the pairwise compatibility term between parts as follows. 

\begin{equation}
D(I,p_{i},p_{j})=\omega_{ij}^{s_{i}s_{j}}\psi{}_{ij}(p_{i},p_{j})+b_{ij}^{s_{i}s_{j}},\label{eq:pairwise}
\end{equation}

where $\psi_{ij}(p_{i},p_{j})=[dx,dy,dx^{2},dy^{2}]$ denotes the
relative distance between $p_{i}$ and $p_{j}$ , $\omega_{ij}^{s_{i}s_{j}}$
denotes the symbol-specific weights, and $b_{ij}^{s_{i}s_{j}}$ denotes
the bias of the compatibility. If two symbols are not ``compatible'',
\textit{i.e.}, they never exist in any training image together, this
bias term is $-\infty$. Both terms are learned in Sec. \ref{sub:Our-model}.

In graphical models, we frequently model the energy minimization problem by passing messages from one
node to the other.
This message passing step is practical in both exact inference or inexact inference.
We can utilize this
compatibility term, which result in fine scale message passing.
For node $P_j$ , the incoming message $m_{k\rightarrow j}(p_{j},s_{j})$ from other nodes $P_k$ and the outgoing message $m_{j}(p_{j},s_{j})$ are computed as:

\begin{equation}
m_{j}(p_{j},s_{j})=\sum_{k\in n(j)}m_{k\rightarrow j}(p_{j},s_{j})\label{eq:3}
\end{equation}

\begin{equation}
m_{k\rightarrow j}(p_{j},s_{j})=\max_{s_{k}}\left[\max_{p_{k}} \left[m_{k}(p_{k},s_{k})+D(I,p_{k},p_{j})\right]\right]\label{eq:4}
\end{equation}

where $n(j)$ denotes the neighbors of $p_{j}$ in $G$.

\subsection{Inference\label{sub:Our-model}}

Given a set of visual symbols, compatibility map, and their graph
structure $G$, one can learn the parameters and perform inference.
When the structure is a tree, the inference is exact.

In our experiment, we define the following compositional parts (Fig.
\ref{fig:vc}d). Semantically, our structure has three levels: ``upper
body'' and ``lower body'' as a coarse modeling of the human body,
head, upper and lower limbs used in midlevel description, and joints in
the fine level.

\paragraph*{Appearance term}

We use HOG templates to represent each visual subcategory.
For each part $p_{i}$ in an image $I$, the appearance score of a local
patch can be written as \cite{yang2011articulated} 
\begin{equation}
B(I,p_{i})=\omega_{i}^{s_{i}}\phi(I,p_{i}),
\end{equation}
where $\omega_{i}^{s_{i}}$ is the weight for symbol $s_{i}$ in the $i^{th}$
part. This term can be initialized by the results $w_k$ in Eq. \ref{eq:1}.

\paragraph*{Deformable term}

Pairwise term between $p_{i},p_{j}$ is defined as the symbol-wise
context (Eq. \ref{eq:pairwise}). This term can be computed effectively
by distance transform in inference.

\paragraph{Objective function}

Our objective function is as follows

\begin{equation}
p=\arg\max_p\sum B(I,p_{i})+\sum_{i,j}D(I,p_{i},p_{j})\label{eq:6}
\end{equation}

Since our model is a tree, standard message passing algorithm (Eq. \ref{eq:3} and Eq. \ref{eq:4}) and exact inference are applicable.

\paragraph{Learning model parameters}

The objective function (Eq. \ref{eq:6}) can be rewritten as $f=\langle\theta,\Phi\rangle$
, where $\langle \cdot,\cdot \rangle$ is the inner product, $\theta$ consists of image filters for single parts ($\omega_i^{s_i}$), pairwise deformable weights ($\omega_{ij}^{s_{i}s_{j}}$) and biases $b_{ij}^{s_i,s_j}$, and $\Phi(I_{n},p)$ denotes the
concatenated features from appearance and deformable components. The
learning of $\theta$ amounts to the quadratic optimization: 
\begin{equation}
\begin{split}\arg\min_{\theta,\xi_{i}\geq0}\frac{1}{2}\theta^T\cdot\theta+C\sum_{n=1}^{N}\xi_{i},\\
\forall n\in\text{pos}\quad\theta^{T}\Phi(I_{n},p)\geq1-\xi_{n},\\
\forall n\in\text{neg}\quad\theta^{T}\Phi(I_{n},p)\leq-1+\xi_{n}.
\end{split}
\label{eq:7}
\end{equation}

This is a standard quadratic programming procedure, and can
be solved effectively.

\section{Experiments}
We present our experiments in this section. First, we describe the
datasets we used for evaluation. Then, we demonstrate the visual symbols
learned by our method, and compare our approach against four other
methods. 

In our experiments, we extract HOG features on grid image with $4\times4$
pixels from image patches, and learn visual symbols and geometric
context map. The number of geometric cluster $k_{j}$ is $8$ for large
compositional parts, and $6$ for small parts, and the number of visual
symbols for each geometric types is set to $2$ and $4$, respectively. 
The number of geometric clusters is consistent with that in \cite{yang2011articulated}.
The final number of the appearance clusters depends on the cross validation.
As a result, we learn 8 to 20 visual categories for each visual symbol after cross validation. 

\subsection{Dataset}
We evaluate our performance on two large datasets, namely, Image Parse
dataset \cite{ramanan2007learning} and Leeds Sport dataset \cite{Johnson10}.
In all experiments, we used 500 images in the negative set of INRIA
person dataset as negative samples (\cite{dalal2005histograms})

\subsubsection*{Image Parse dataset}
Image Parse dataset (PARSE) contains 305 images with annotated poses.
This dataset has images from various human activities, background
and different illuminations. All images are resized such that human
in images have roughly the same scale (150 pixels in height). We used
100 images for training, and the rest for testing.

\subsubsection*{LSP dataset}
The recent Leeds Sport Dataset (LSP) contains 2000 images. This collection
has a larger variation of pose changes. Humans in each image were
cropped and scaled to 100 to 150 pixels in height. The dataset was
split evenly into training set and testing set.
\begin{figure}
\begin{centering}
\includegraphics[width=1\columnwidth]{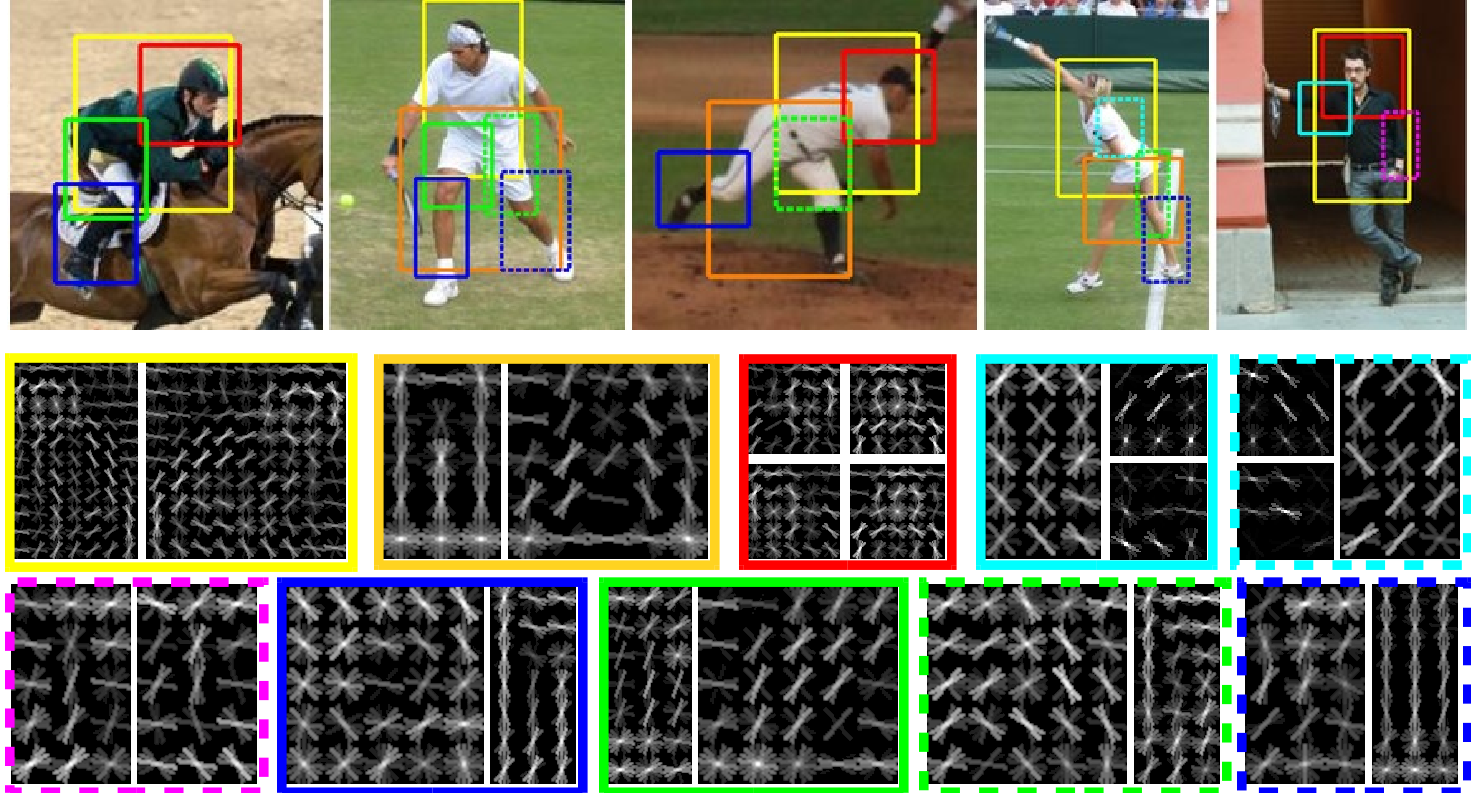} 
\par\end{centering}

\caption{\label{fig:demo}We show filters for our compositional parts: torso
(yellow), lower body (orange), head (red), upper arm (cyan), lower
arm (magenta), upper leg (green), lower leg (blue). Left/right side
are denoted by solid/dashed lines, respectively.}
\end{figure}

\subsection{Demonstration}

We demonstrate the effectiveness of learning procedure in this section.
Fig. \ref{fig:demo} shows localization results for the eleven higher
level compositional parts, as well as examples of their filters learned
by our method. In this visualization, we use different colors and
line types to denote different parts.

\paragraph*{Filters for visual symbols}

Each filter in Fig. \ref{fig:demo} exhibits a few characteristics
for the corresponding compositional part. These filters are related,
in the sense that they model the human body at different levels. Each filter is also self-contained,\textit{ e.g.}, any one is not
the sub-region of another due to the training process. This intrinsic
constraint facilitates the inference. For instance, the torso (yellow)
filters indicate the body inclination in a coarse level, which limits
the search of head position (red) both in geometric context and appearance.

\paragraph*{Interpretation}

The localization results can be regarded as multi level symbolic annotations
for the human body. Therefore, this can be used for a number of applications
ranging from high level understanding to low level description. We
can further assign semantic meaning to these symbols. For instance,
the torso and lower body locations can be used for analyzing the actions
of the human beings (``stretching'', ``standing'', ``squat'',
etc), and the midlevel limb detection results are used for motion
analysis (``extending arm'', ``fetching'', etc.).

\begin{figure*}[!t]
\begin{centering}
\includegraphics[width=1\textwidth]{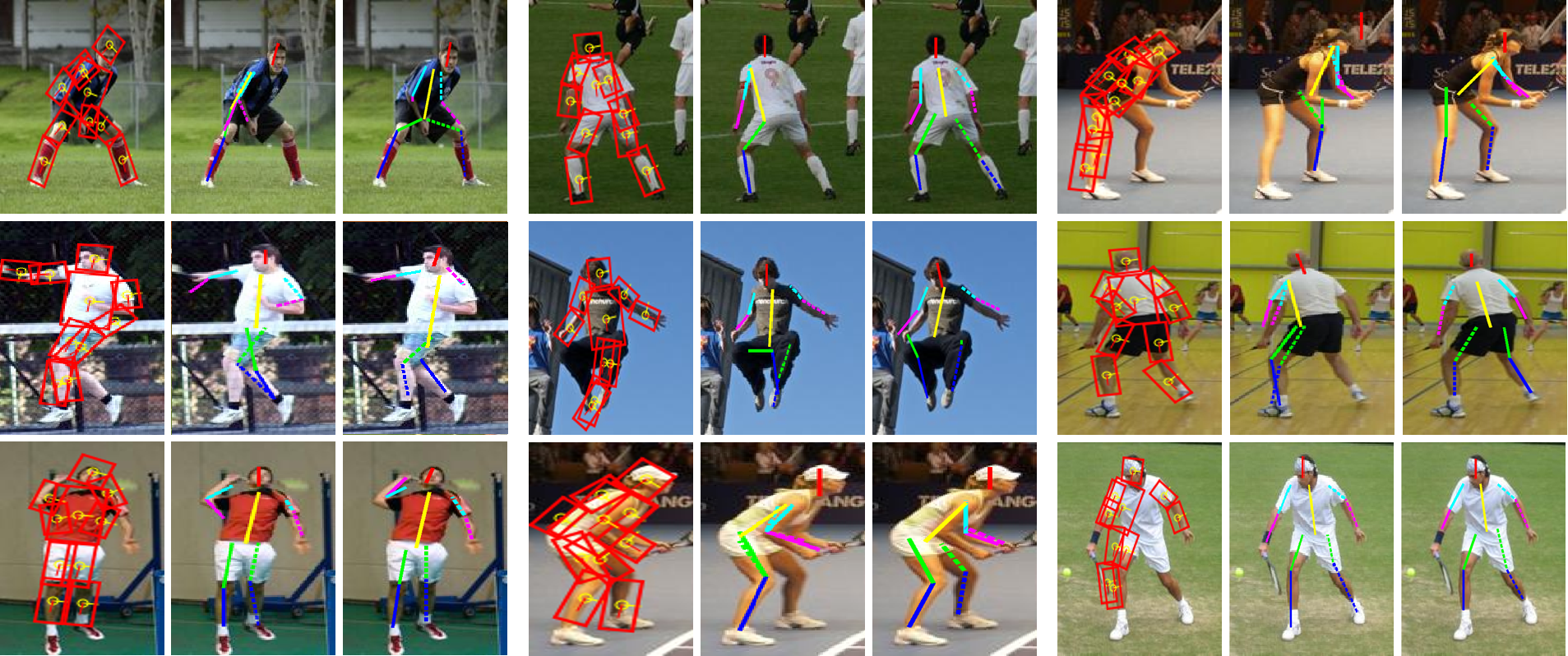}
\par\end{centering}

\caption{Result comparison. Each triplet of image contains results for {[}Andriluka
\textit{et al.}, 2009{]} in its original visualization, {[}Yang and
Ramanan, 2011{]} and ours using the visualization protocol in {[}Johnson
and Everingham, 2011{]} . \label{fig:Visual-comparison}}

\end{figure*}

\begin{table*}[!t]
\begin{centering}
\begin{tabular}{|c|c|c|c|c|c|c|c|c|}
\hline 
Exp.  & Method  & Torso  & Head  & Upper Leg  & Lower Leg  & U.Arm  & L.Arm  & Total \tabularnewline
\hline 
\hline 
 PARSE  & \cite{yang2011articulated}  & \textbf{97.6}  & \textbf{93.2 } & 83.9  & 75.1  & 72.0  & \textbf{48.3}  & 74.9 \tabularnewline
\hline 
 & Ours  & 97.1 & 90.2  & \textbf{86.1 } & \textbf{77.1 } & \textbf{74.9 } & 46.9 & \textbf{75.7} \tabularnewline
\hline 
\hline 
LSP  & \cite{yang2011articulated}  & 92.6  & 87.4  & 66.4  & 57.7  & 50.0  & 30.4  & 58.9 \tabularnewline
\hline 
 & \cite{DBLP:conf/eccv/TianZN12} (first 200)  & 93.7  & 86.5  & 68.0  & 57.8  & 49.0  & 29.2  & 58.8 \tabularnewline
\hline 
 & \cite{DBLP:conf/eccv/TianZN12} (5 models)  & \textbf{95.8}  & \textbf{87.8}  & 69.9  & 60.0  & 51.9  & 32.9  & 61.3 \tabularnewline
\hline 
 & \cite{Johnson10}  & 78.1  & 62.9  & 65.8  & 58.8  & 47.4  & 32.9  & 55.1 \tabularnewline
\hline 
 & \cite{DBLP:conf/cvpr/JohnsonE11}  & 88.1  & 74.6  & 74.5  & 66.5  & 53.7  &  \textbf{37.5}  & 62.7\tabularnewline
\hline 
 & \cite{andriluka2009pictorial}  & 76.8  & 68.5  & 56.9  & 48.8  & 37.4  & 20.0  & 47.1 \tabularnewline
\hline 
 & Ours  & 92.2  & 84.7  &  \textbf{78.1}  & \textbf{67.5}  & \textbf{54.7}  & 37.2  & \textbf{65.2}\tabularnewline
\hline 
\end{tabular}
\par\end{centering}

\caption{Performance on the PARSE and the LSP dataset. The first two rows shows
the performance comparison on the PARSE dataset against {[}Yang and
Ramanan, 2011{]}. The next seven rows show the performance of five
algorithms on the more challenging LSP dataset.\label{tab:Comparison-results.}}
\end{table*}

\subsection{Comparison}

We compared our approach against four state of the art methods for
human poses parsing on the PARSE and the LSP dataset in this section.
In our comparison, we used the criterion in \cite{DBLP:conf/cvpr/FerrariMZ08}
for performance evaluation. A part is correctly detected if both its
endpoints are within $50\%$ of the length of corresponding ground
truth segments. Then we used the probability of a correct pose (PCP)
to measure the percentage of correctly localized body parts.

Table \ref{tab:Comparison-results.} summarizes the evaluation results,
with highest scores being highlighted. We compared the parsing accuracy
of our method against \cite{andriluka2009pictorial}, \cite{yang2011articulated}, \cite{Johnson10}, and \cite{DBLP:conf/eccv/TianZN12},
respectively. We re-run the method of \cite{andriluka2009pictorial}
and \cite{yang2011articulated} on the datasets and report the results,
and other data entries in the table are from the original papers,
respectively. Please note that \cite{DBLP:conf/eccv/TianZN12} tried
two different settings in their methods, but they did not report the
result using all the 1000 images in the training set. 

Overall our method achieved the best performance. In the PARSE dataset, our method is marginally better
than the original algorithm ($75.7\%$ vs $74.9\%$ ). This is possibly
because the power of our visual category training may not be fully
explored due to small number of training samples. 

The problem caused by limited training samples is relieved in the
LSP dataset. When 1000 images are used for training, our method
outperforms other methods. 
Compared to four methods, our total detection accuracy ($65.2\%$) is consistently higher.
Our performance is also superior to \cite{DBLP:conf/cvpr/JohnsonE11},
where 11000 samples were used for training%
\footnote{Due to a large number of missing labels in this dataset, we do not
perform our evaluation on it. In their method, the training samples were automatically relabeled during optimization.
}. 
This suggests that our training is effective.

Fig. \ref{fig:Visual-comparison} shows some examples of parsing results
for three methods. Each triplet contains results for \cite{andriluka2009pictorial},
\cite{yang2011articulated} and ours. The parsing results show that
our method produces visually pleasing results. The interaction between
high level and low level compositional parts makes our results more
``balanced''. For instance, detection results for two legs are well
separated if the corresponding symbols of the lower body part are
detected, because the symbol-wise geometric context naturally guides
the maximization (Eq. \ref{eq:6}) to this optimal solution. Our method
also tries to make the best guess of self-occluded parts.

The training takes approximately 8 hours on a 2.8GHz Quad Core CPU with 6GB memory. 
Test takes 1.5s for an $320\times240$ image. 
Compared to models where loopy BP is used, our tree structure essentially speeds up the training.
The running time of our method is in the same order of magnitude of
\cite{yang2011articulated}. Therefore, our method strikes a balance
between accuracy from long range interaction and the efficiency from
exact inference.

\section{Conclusion}

This paper presents a novel approach to learning self-contained representations
for parsing human poses in images. The main contribution is the visual
symbols that facilitate geometric context modeling. Our method can
be used for many graphical models. When the model is a tree, we demonstrate
that our method outperforms four current methods.

\bibliographystyle{named}
\bibliography{ijcai13}

\end{document}